\documentclass{ieeeaccess}
\usepackage{cite}
\usepackage{amsmath,amssymb,amsfonts}
\usepackage{graphicx}
\usepackage{textcomp}
\usepackage{makecell}
\usepackage{tabu}
\usepackage{algorithm}
\usepackage{algorithmicx, algpseudocode}

\def\BibTeX{{\rm B\kern-.05em{\sc i\kern-.025em b}\kern-.08em
    T\kern-.1667em\lower.7ex\hbox{E}\kern-.125emX}}
\begin{document}
\history{Date of publication xxxx 00, 0000, date of current version xxxx 00, 0000.}
\doi{10.1109/ACCESS.2017.DOI}
\title{Effects of Sim2Real Image Translation on Lane Keeping Assist System in CARLA Simulator}
\author{\uppercase{Jinu Pahk}\authorrefmark{*}\authorrefmark{1},
\uppercase{Jungseok Shim\authorrefmark{*}\authorrefmark{1}, MinHyeok Baek\authorrefmark{1}, Youngseob Lim\authorrefmark{2}, and Gyeungho Choi\authorrefmark{3}}}
\address[1]{Daegu Gyeongbuk Institute of Science and Technology, Daegu 42988, South Korea (e-mail: {kyg132, shim2751, minheak06}@dgist.ac.kr)}
\address[2]{Department of Robotics Engineering, Daegu Gyeongbuk Institute of Science and Technology, Daegu 42988, South Korea}
\address[3]{Department of Interdisciplinary Engineering, Daegu Gyeognbuk Institute of Science and Technology, Daegu 42988, South Korea}

\markboth
{J. Pahk et al. \headeretal: Effects of Sim2Real Image Translation on Lane Keeping Assist System in CARLA Simulator}
{J. Pahk et al. \headeretal: Effects of Sim2Real Image Translation on Lane Keeping Assist System in CARLA Simulator}

\corresp{Corresponding author:  Yongseob Lim (e-mail: yslim73@dgist.ac.kr) and Gyeungho Choi (e-mail: ghchoi@dgist.ac.kr).\\ *JINU PAHK and JUNGSEOK SHIM are co-first authors.}

\begin{abstract}
Autonomous vehicle simulation has the advantage of testing algorithms in various environment variables and scenarios without wasting time and resources, however, there is a visual gap with the real-world. In this paper, we trained DCLGAN to realistically convert the image of the CARLA simulator and evaluated the effect of the Sim2Real conversion focusing on the LKAS (Lane Keeping Assist System) algorithm. In order to avoid the case where the lane is translated distortedly by DCLGAN, we found the optimal training hyperparameter using FSIM (feature-similarity). After training, we built a system that connected the DCLGAN model with CARLA and AV in real-time. Then, we collected data (e.g. images, GPS) and analyzed them using the following four methods. First, image reality was measured with FID, which we verified quantitatively reflects the lane characteristics. CARLA images that passed through DCLGAN had smaller FID values than the original images. Second, lane segmentation accuracy through ENet-SAD was improved by DCLGAN. Third, in the curved route, the case of using DCLGAN drove closer to the center of the lane and had a high success rate. Lastly, in the straight route, DCLGAN improved lane restoring ability after deviating from the center of the lane as much as in reality.
\end{abstract}


\titlepgskip=-15pt

\maketitle

\section{Introduction}
\newcommand\tab[1][0.25cm]{\hspace*{#1}}
\label{sec:introduction}\PARstart{C}{urrently,} the autonomous driving industry is expanding at a rapid pace. Autonomous driving has been proposed as a solution to traffic accidents that cause 1.25 million deaths worldwide every year \cite{b1}. However, in the United States alone, a total of 392 traffic accidents occurred from July 2021 to May 2022 due to Level 2 autonomous driving, and these reports raised the need for securing the safety of autonomous driving technology \cite{b2}.\\ \tab Autonomous driving simulation is drawing attention as a new testbed. Unlike reality, the simulation can implement complex situations, such as traffic accidents and specific weather, that are hard to reproduce in the real-world \cite{b3, b4, b5}. This technology gives train datasets about abnormal situations and will prevent traffic accidents more perfectly. Therefore, many studies have been conducted for modeling vehicles in simulation \cite{b6, b7, b8}. \Figure[!t](topskip=0pt, botskip=0pt, midskip=0pt)[height=1.75in]{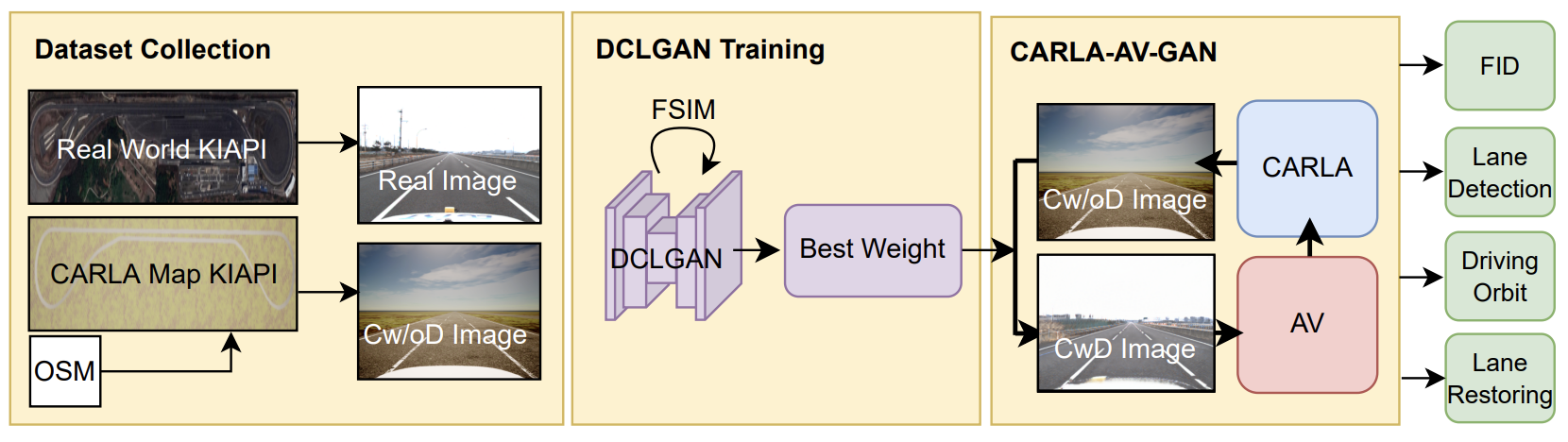}{ Full framework for identifying Sim2Real image gaps and driving gaps from the CARLA-AV-GAN loop.\label{fig1}}\\ \tab In order to adapt proven algorithms from virtual-world to real-world, reducing the gap between simulation and reality is an inevitable task. This gap is caused by several factors such as the unreality of sensor models (e.g., camera, LiDAR, GPS, and IMU), mismatch of dynamic models, and noise in the real environment. Among the factors, the camera sensor’s input images are the main cause of the gap because this sensor accounts for the largest portion of driving. The image's gap which is a difference in visual realism between the real domain and the simulation domain would make driving results unreliable. Moreover, correcting this gap only by adjusting the simulation parameters is impossible. Therefore, simulation developers and users often have used a high-level graphic engine or imported a deep learning model to make the visual contents of the simulation look more realistic.\\ \tab To make images from the two domains similar, Generative Adversarial Networks (GAN) \cite{b9} has been adopted. With the development of the GAN structure, the model has begun to be applied to various fields such as colorizing gray image \cite{b10} and restoring image quality \cite{b11}. Moreover, many studies have attempted to make virtual world images similar to reality \cite{b12, b13}. Although the study on improving the graphic quality of the virtual world is active, autonomous driving tests with these photorealistic techniques have not been dealt with yet.
\\ \tab The purpose of this paper is to show the possibility of applying GAN to the autonomous vehicle (AV) simulation not just translating images, but also driving tests. We used CARLA \cite{b14} for AV simulation. Dual Contrastive Learning Generative Adversarial Networks (DCLGAN) \cite{b15} was applied because DCLGAN showed the better performance to translate lane geography characteristics than CycleGAN \cite{b16} without mode collapse. Moreover, we explored a way to train DCLGAN using FSIM to prevent lane distortion.
\\ \tab For this purpose, the data were collected from the three domains: real-world, CARLA with DCLGAN (hereafter referred to as CwD), and CARLA without DCLGAN (hereafter referred to as Cw/oD). The real-world data were obtained from the Korea Intelligent Automotive Parts Promotion Institute (KIAPI) and the virtual-world data were obtained from a map the same as KIAPI. The types of data were camera images and GPS data. To show the DCLGAN contributed to making more realistic driving results, a method to quantify the Sim2Real gap is needed. We propose the following four methods: First, FID(Fréchet Inception distance) \cite{b17} measures the reality of the virtual-world's images by calculating differences in density of two distributions of imageset. Second, lane segmentation performance was measured and compared. Third, we calculated the driving orbital difference between the real-world and the simulation with/without DCLGAN by using GPS data. Finally, we measured how well the vehicle returns to its original lane when it is out of the center. Experimental results show that the performance of our proposed CwD method is more effective in practical use than without DCLGAN.

\section{Related Works}

\subsection{CARLA}
\tab CARLA is an autonomous driving research simulator. We used the CARLA simulator for the implementation of a vehicle and conducting environments by using the following functions. First, ROS-bridge was provided. ROS-bridge enables the connection between real-vehicle and virtual-vehicle. Second, CARLA allowed us to spawn various sensors and configure specific parameters such as the camera’s FoV, resolution, and gamma. Lastly, custom vehicles can be implemented through Unreal Engine 4 (UE4), the game engine of CARLA. We imported a 3D model of the same size as the Hyundai IONIQ Electric for spawn sensors in the same position as reality.

\subsection{Generative Adversarial Network}
\tab Generative Adversarial Network consists of a generator that creates fake images following real images’ distribution, and a discriminator that distinguishes between the fake images and the real images. As the two modules compete repeatedly, GAN can generate more realistic images. However, the GAN model has a limitation that it necessarily requires paired datasets for learning. CycleGAN solves the problem by adding the reverse mapping that returns the fake image to the original image. Whereas cycle consistency loss has a drawback that the fake images have low diversity. Contrastive Learning for Unpaired Image-to-Image Translation (CUT) and Dual Contrastive Learning GAN tried to deal with this issue through continuous learning. Especially, Similarity Dual Contrastive Learning GAN (SIMDCL) with similarity loss added to DCLGAN suffers less from the mode collapse by allocating embeddings as much as the number of data domains. We mainly used SIMDCL, considering the characteristics of our dataset, which has low diversity and a high risk of mode collapse. 

\subsection{Fréchet Inception Distance}
\tab FID  is an evaluation metric for a generative model such as Generative Adversarial Networks (GAN), text-to-image, etc. The distance between the fake images and the real images is measured by comparing the extracted features from the pre-trained Inception v3 model with  the Imagenet dataset. In this paper, the measured distance presented how realistic the generative model creates an image. FID is calculated as \eqref{eq1} using the mean of the variance of features ($\mu$) and covariance($\sum$).

\begin{equation}FID = \left | \mu _T - \mu_G \right | ^ 2 + Tr(\sum _T + \sum_F-2(\sum_T\sum_G)^\frac{1}{2}) .\label{eq1}\end{equation}

\subsection{ENet-SAD}
\tab ENet-SAD \cite{b18} has the characteristic of self-learning, which delivers and enhances the representation top-down by distilling the attention map of the lower layer. Thanks to this distillation technique, Enet-SAD can be computed faster without additional annotation or inference even with much fewer parameters compared to other CNN-based lane detection models such as SCNN. Since the IoU of the ground truth lane and lane segmentation prediction results were better than the previous model, our LKAS system uses ENet-SAD.

\subsection{TuSimple Evaluation}
\tab Tusimple \cite{b19} introduces the lane segmentation accuracy formula as \eqref{eq2}. C$_{clip}$ is the number of true points in each clip and S$_{clip}$ is the number of the ground truth points in the clip. We used this evaluation method to compare each domain's accuracy.

\begin{equation}Accuracy = \frac{\sum_{clip}C_{clip}}{\sum_{clip}S_{clip}}.\label{eq2}\end{equation}

\section {Environment Setting}
\subsection{Real-world Setting}
\subsubsection{Specification}
\tab TABLE \ref{table1} describes the sensors and hardware’s specifications. Fig. \ref{fig2} shows the location of the sensor. Although the camera’s maximum resolution is 1632x1248 pixels, 808x620 was used by binning. The vehicle was the Hyundai IONIQ Electric modified for applying the drive-by-wire system.\Figure[h](topskip=0pt, botskip=0pt, midskip=0pt)[height=1in]{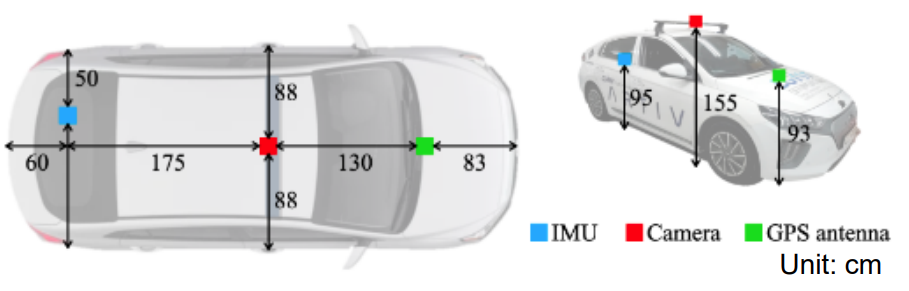}{Vehicle top view and sensor location (left). Sensor installation on the test vehicle (right).\label{fig2}} 

\begin{table}[h]
\setlength{\tabcolsep}{5pt}
{\tabulinesep=0.5mm
\begin{tabu}{cccc}
\hline
\makecell[c]{Hardware\\Component}& Purpose& Model& Specifications \\
\hline \hline
\makecell[c]{Main \\PC}&\makecell[c]{ Vehicle \\Control} &\makecell[c]{Nuvo\\-8108GC\\-NX}&
\makecell[c]{CPU:\\ Intel Xeon Processor \\E-2176G  \\Memory:\\ 32GB DDR4 \\GPU: \\GeForce RTX-2080Ti\\ 11GB}\\
\hline
\makecell[c]{Deep \\Learning \\PC}&  \makecell[c]{Running \\deep learning \\algorithm}& \makecell[c]{Custom\\PC}& 
\makecell[c]{CPU: \\Intel I7 9700F \\Memory: \\32GB DDR4\\
GPU: \\NVIDIA Titan X\\12GB} \\
\hline
\makecell[c]{Simulation \\PC}& \makecell[c]{Playing \\CARLA \\ Calculating \\GAN model}& \makecell[c]{Custom\\ PC}&
\makecell[c]{CPU:\\ AMD Ryzen 9 3950x\\Memory:\\ 64GB DDR4 \\ GPU: \\GeForce RTX-3090\\24GB}\\
\hline
\makecell[c]{FLIR \\Camera}&\makecell[c]{Lane \\recognition}&\makecell[c]{BFS-PGE\\-19S4C-C}& \makecell[c]{FoV: 76$^{\circ}$ \\FPS: 50 \\Gamma: 0.8\\
Resolution:\\ 1632x1248 \\(808x620 after binning)}\\
\hline
\makecell[c]{GPS\\(RTK GNSS)}&\makecell[c]{Checking \\the vehicle \\trajectory}&MRP-2000&
\makecell[c]{RTK Accuracy: \\0.010m + 1ppm CEP \\ Rate: up to 10Hz}\\
\hline
\end{tabu}}
\caption{Sensors and Hardware’s specification.}
\label{table1}
\end{table}

\subsubsection{LKAS algorithm}
\tab We have Autonomous driving algorithms and architecture that were licensed by MOLIT (Ministry of Land, Infrastructure, and Transport) of the Republic of Korea. This license required Autonomous Emergency Braking, Smart Cruise Control, and Lane Keeping Assist (LKA). Among the technologies we have, only LKA was used in the experiment.\\ \tab Previous paper\cite{b21, b22} proposed an lane keeping algorithm. This algorithm worked in two stages. The first stage was lane segmentation. In Deep Learning PC, Enet-SAD was used to segment the lane. This Enet-SAD was trained by datasets from the CULANE Dataset\cite{b23} and our laboratory. The second stage was lane detection. The lane detection transformed the segmented image to a bird's eye view, then calculated lane center points. After these stages, Main PC adjusts steer and velocity by using pure pursuit through the result of lane detection. In this paper, lane keeping was performed in KIAPI and compared with the data from the actual vehicle and the virtual vehicle.

\subsubsection{Proving Ground}

\tab The real-world test data was collected at the Korea Intelligent Automotive parts Promotion Institute (KIAPI), located in Daegu, Korea. Among the various test circuits, the high-speed circuit was used. The high-speed circuit is a telephone receiver-shaped and consists of a three-lane straight line and bank. Fig. \ref{fig3} shows the actual appearance of the facility.
\Figure[h](topskip=0pt, botskip=0pt, midskip=0pt)[width=3.3in]{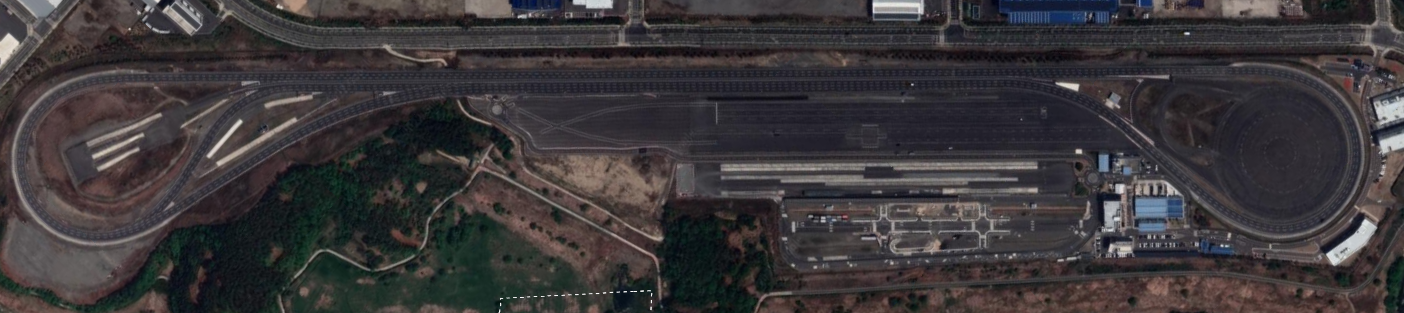}{KIAPI satellite picture.\label{fig3}} 

\subsection{CARLA Setting}
\subsubsection{Camera and Vehicle}
\tab CARLA provides various sensors and attributes. Attributes make it possible to set the sensor's specifications. The performance of the camera mentioned in TABLE \ref{table1} was imported such as FoV or Gamma. Then, the camera was spawned according to Fig. \ref{fig2} position. The CARLA image received from the imported camera and the real-world image can be seen in Fig. \ref{fig4}.
\Figure[h](topskip=0pt, botskip=0pt, midskip=0pt)[width=3.3in]{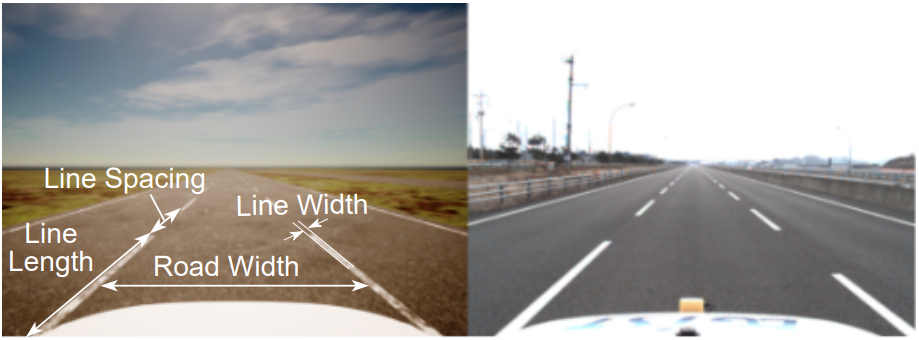}{CARLA image (left) and Real-world image (right).\label{fig4}}

\subsubsection{Map}
\tab To make a custom map, CARLA needed map geometry information in .fbx format and OpenDRIVE information in .xodr format. For these, We used RoadRunner recommended by CARLA as a map editor and OpenStreetMap (OSM) as map data. OSM has latitude and longitude as vector data and RoadRunner made OSM possible to be map geometry information and OpenDRIVE information.\\

\begin{table}[h]
\setlength{\tabcolsep}{28pt}
{\tabulinesep=0.5mm
\begin{tabu}{cc}
\hline
Parameters & Specification\\
\hline
\hline
Road Width (m)& 3.5\\
\hline
Line Length (m) & 4.5\\
\hline
Line Spacing (m) & 4\\
\hline
Line Width (m) &0.125\\
\hline
Slope ($^\circ$) &\makecell[c]{30 (first lane)\\0 (second, third lane)}\\
\hline
\end{tabu}}
\caption{Road parameters' specification.}
\label{table2}
\end{table}

\tab The process of making the KIAPI map was as follows. First, as can be seen in Fig. \ref{fig5}(a) we dragged the desired area from the satellite map in OSM and export the file in .osm format. Second, as shown in Fig. \ref{fig5}(b), the latitude and longitude were corrected by comparing the osm file exported from JOSM with the GPS file of actual measurement data. After that, the calibrated osm file was imported into RoadRunner, and the RoadRunner file set the road width, lanes, and slope of the road according to the information announced in KIAPI \cite{b24}. Fig. \ref{fig5}(c) describes this process and TABLE \ref{table2} shows the specification of each parameter. Lastly, this file was exported  in Carla format such as fbx and xodr format files.

\Figure[h](topskip=0pt, botskip=0pt, midskip=0pt)[width=2.0in]{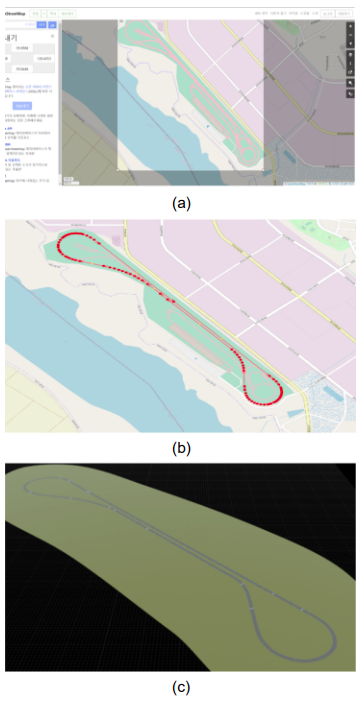}{The process of creating a virtual map from the OSM data of KIAPI. (a) Exporting map data from OSM. (b) PG osm file edited on JOSM. (c) KIAPI map produced in RoadRunner.\label{fig5}} 

\subsubsection{Time}
\tab The experiment with the same time period as the real-world is impossible because of the calculation time of GAN. As a solution, we used synchronous time. According to CARLA documentation, “On synchronous mode, the server waits for a client tick, a "ready to go" message, before updating to the following simulation step” [25, Sec.2]. It means that We can set the attribute to wait for a step until the vehicle control command arrived. Therefore, the Client server would wait for a cycle described in 3.B.3. section.

\subsubsection{Network Connection between CARLA and AV}
\tab Three computers were used for the experiments, each called the Main PC, Deep Learning PC (DL PC), and Simulation PC. TABLE \ref{table1} shows specification. Main PC decided on vehicle control such as steering value and accelerator value. DL PC computed deep learning algorithms such as the LKA algorithm. Simulation PC played CARLA simulation. 10 Gbps cables were used to connect three PCs. We selected ROS as a middleware. TABLE \ref{table3} describes the ROS version for each computer. For connecting different versions of ROS 2 between DL PC and Simulation PC, Cyclone DDS is applied to Simulation PC and  Fast RTPS is used to DL PC.
\begin{table}[h]
\setlength{\tabcolsep}{20pt}
{\tabulinesep=3mm
\begin{tabu}{ccc}
\hline
PC Name & Ubuntu Ver. & ROS Ver.\\
\hline\hline
Main PC & 18.04 LTS & Melodic \\
\hline
Deep Learning PC & 18.04 LTS & Dashing \\
\hline
Simulation PC & 20.04 LTS & Foxy \\
\hline
\end{tabu}}
\caption{Each PC’s Ubuntu version and ROS version.}
\label{table3}
\end{table}
 \\ \tab Fig. \ref{fig6} shows flow of the three computers interactions. The title of the box means the topic name. First, in the Simulation PC, the camera sensor from CARLA gave an image to GAN. The output from the GAN model was sent to the DL PC.  Second, in the DL PC, lane\_segmentation function was performed and gave segmentation data to lane\_detect function. The lane\_detect function found center points of the lane and sent the output points to the Main PC. Third, in the Main PC, lane\_driving function got vehicle status from the Simulation PC and center points from DL PC. By utilizing these values, the lane\_driving function determined desired steer value. move\_car function calculates proper steer, acceleration, and brake values. Then the function sent them to the Simulation PC for control of the vehicle. 
\Figure[h!](topskip=0pt, botskip=0pt, midskip=0pt)[width=3.3in]{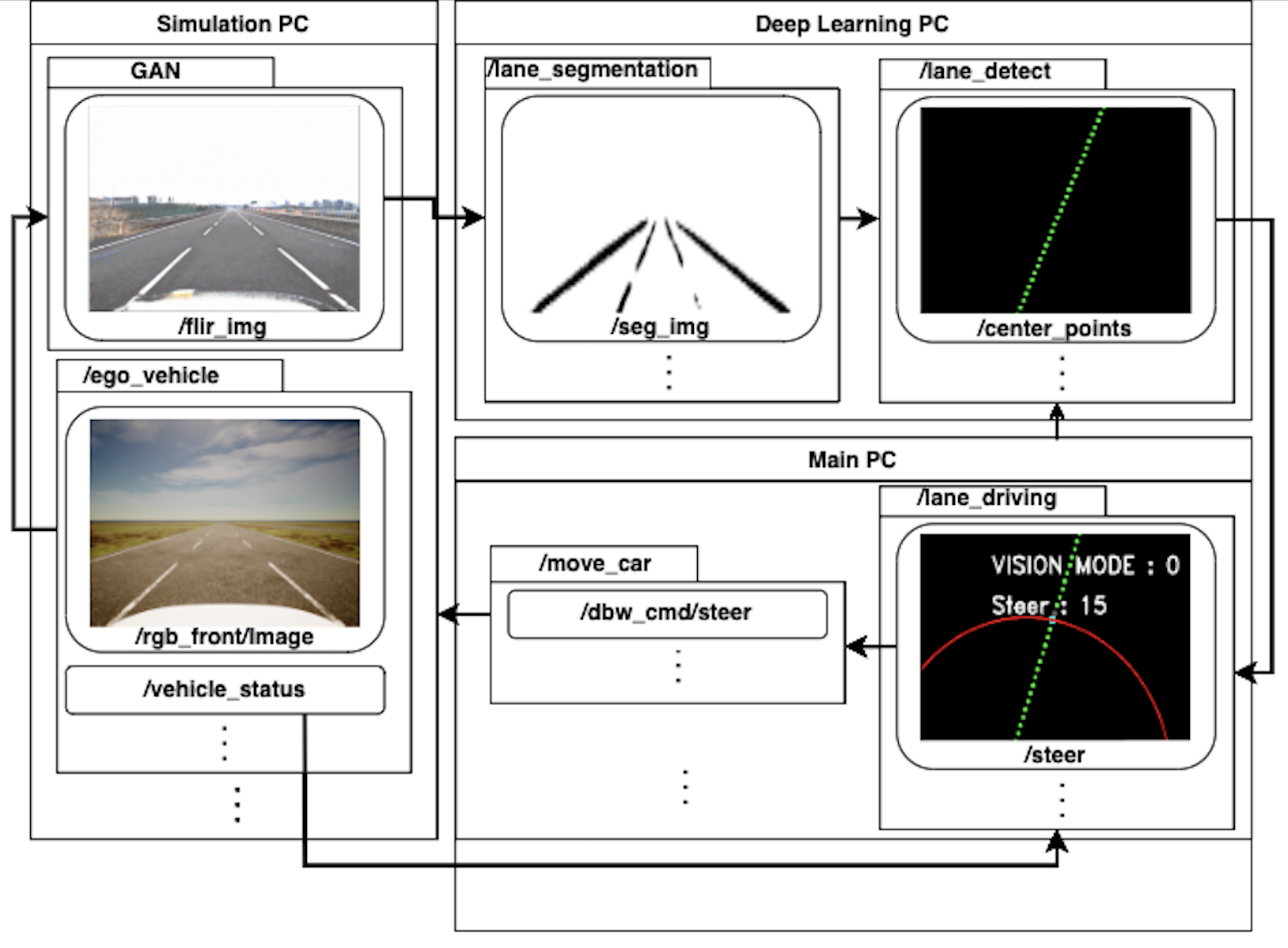}{ CARLA-AV-GAN Connection overview.\label{fig6}} 

\section{Approach Method}
\subsection{Lane Characteristics Difference Measurement using FID}

\tab It is necessary to guarantee whether the difference in characteristics of lanes can be quantitatively measured. As shown in Fig. \ref{fig4}, the lanes in the real-world are clearer and thicker. To check whether the FID used to quantify the Sim2Real difference in this paper can identify the difference between these lanes, we made several maps. These maps were based on KIAPI map and varied the thickness, texture, length, and distance of lanes using RoadRunner. Fig. \ref{fig7} shows these characteristic differences. For the texture, LaneMarking1, 2, and 3, which are lane materials provided by RoadRunner, were used. The material used in LaneMarking1 is the sharpest and 3 is the faintest. All other variables in RoadRunner were fixed except for lanes.
\\\tab After that, the vehicle was summoned using the RoadRunner Scenario function and then driven on the KIAPI map. While the car was driving in the center of the road, 25-second videos were recorded on each map with specific characteristics of the lane. By extracting each frame of the recorded video, about 700 imageset was created for each map, and FID operation was performed between the maps using this. At this time, the FID code was written based on PyTorch and the dimensionality of features was set to 2048 (default) \cite{b26}.  
\\\tab To compare only the thickness of lanes, we set the lane material to LaneMarking1, set the length and spacing of lanes to (10m, 10m), and increased the lane thickness from 0.125m to 0.15m, 0.175m, 0.2m. While going, the FID between each other was measured. Also, to compare the texture, the thickness was fixed to 0.125m and the length and spacing of the lanes were fixed to (10m, 10m). To compare the length and spacing of the lanes, the thickness was fixed to 0.125m and the material to LaneMarking1.

\Figure[h](topskip=0pt, botskip=0pt, midskip=0pt)[width=3.3in]{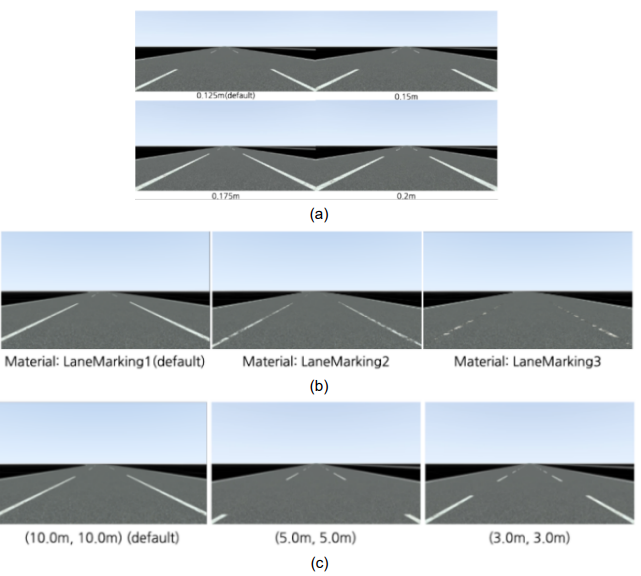}{  Front camera image in RoadRunner Simulation Play according to (a) the thickness of lanes, (b) the material of lane texture, (c) the length of the lane and the distance between lanes. \label{fig7}}

\tab TABLE \ref{table4} shows the FID measurements between each map. The fact that the value increases toward the upper right of the table mean that the FID actually appears larger as the characteristics of the lanes are quantitatively farther from each other. For example, the thickness change result (TABLE 3 (a)) shows that the FID is smaller when the thickness of the lane is quantitatively close, and vice versa. When compared with the image of the lane with a thickness of 0.125 m, it can be seen that the FID increases from 3.4 (0.15 m) to 7.6 (0.175m) to 11.2 (0.2 m). For the texture of the lane (TABLE 3(b)), the imageset with the clearest lane (LaneMarking1) was farther from the imageset with the most blurry lane (LaneMarking3) than the distance from the imageset with the second clearest lane (LaneMarking2). (85.1 vs 31.6) The same results (TABLE 3(c)) were obtained for the length and spacing of the lanes. This is thought to be because the Inception v3 model abstracts the characteristics such as thickness, texture, length, and spacing of lanes well and extracts them as features. This tendency confirms that FID is suitable as a numerical value to quantitatively measure the difference in characteristics of the lanes.\\

\begin{table}[h]
\setlength{\tabcolsep}{12pt}
{\tabulinesep=2mm
\begin{tabu}{c||cccc}
\hline
Thickness & 0.125m & 0.15m & 0.175m & 0.2m\\
\hline\hline
0.125m & 0 & 3.36& 7.57 & 11.17\\
\hline
0.15m &  & 0 & 4.18 & 6.71 \\
\hline
0.175m &  &  & 0 & 3.12\\
\hline
0.2m &  &  &  & 0 \\
\hline
\end{tabu}}
\\\begin{center}(a)\end{center}
\setlength{\tabcolsep}{12pt}
{\tabulinesep=1.5mm
\begin{tabu}{c||cccc}
\hline
Texture & \makecell[c]{Lane\\Marking1} & \makecell[c]{Lane\\Marking2} & \makecell[c]{Lane\\Marking3} \\
\hline\hline
LaneMarking1 & 0 & 31.60& 85.14 \\
\hline
LaneMarking2 &  & 0 & 44.05 \\
\hline
LaneMarking3 &  &  & 0 \\
\hline
\end{tabu}}
\\\begin{center}(b)\end{center}
\setlength{\tabcolsep}{11pt}
{\tabulinesep=2mm
\begin{tabu}{c||cccc}
\hline
Length, Spacing& (10m, 10m) & (5m, 5m) & (3m, 3m) \\
\hline\hline
(10m, 10m) & 0 & 15.54& 31.91 \\
\hline
(5m, 5m) &  & 0 & 11.09 \\
\hline
(3m, 3m) &  &  & 0 \\
\hline
\end{tabu}}
\\\begin{center}(c)\end{center}
\caption{FID measurement value between imagesets with (a) different thicknesses of lanes, (b) the material of texture of lanes (c) the lane length and space between lanes.}
\label{table4}
\end{table}

\subsection{Training DCLGAN}

\subsubsection{Training Method}
\tab We tried to reduce this Sim2Real gap measured by FID with DCLGAN. The data of the real-world domain was collected from the high-speed circuit of KIAPI and the data of the CARLA domain was collected from the OSM map. Both were collected while driving at a speed of 30 km/h in the second lane.
\\ \tab The CARLA domain contains 5498 images as trainsets and 611 images as testsets, and the real-world domain contains 6929 images as train set and 770 images as the test set. In order to match the channels of the FLIR camera images used in the AV, the order of the CARLA image channels was changed from RGB to BGR.
\\ \tab The images used for training were loaded with a size of 600x600 and cropped to a size of 300x300. The model was selected as SIMDCL and the rest of the options kept the DCLGAN default settings. Additionally, $\lambda_{NCE,X}$, a hyperparameter of the loss function, was determined through the process of 2).

\subsubsection{Determination of Hyperparameter using FSIM}

\tab As seen in Fig. \ref{fig8}, if the image passes through the DCLGAN model, the lane in CARLA may not maintain its original shape and may be distorted enough to affect lane-keeping driving. This lane distortion occurs when the generator of the GAN does not fully learn the data distribution and is fixed only in a specific biased direction to generate an image\cite{b15}. Also, the imbalance between the straight line data and curve data of the lane can be the cause.\\

\Figure[h](topskip=0pt, botskip=0pt, midskip=0pt)[width=1.5in]{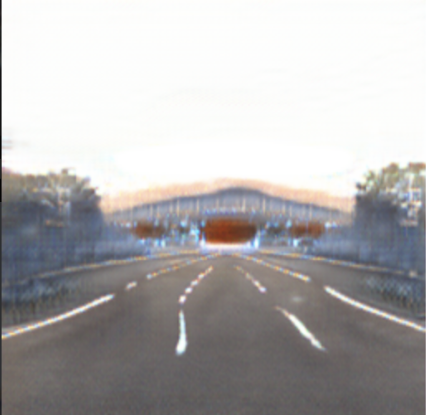}{ Mode collapse occurred when training with the option of DCLGAN model as default. Lane is distorted and all images are similar.  \label{fig8}} 

\tab To solve the problem of distortion of the suboptimal shape, we adjusted the coefficients of the loss function of the generative model. \eqref{eq3} shows the loss function of DCLGAN, and \eqref{eq4} shows the loss function of SimDCL \cite{b11}. As in \eqref{eq5}, the X part of the PatchNCEloss term is associated with a generator that converts X(CARLA)→Y(Real-world). PatchNCELoss calculates Noise Contrastive Estimation(NCE) Loss by receiving the positive pair patch corresponding to query, negative patches, and   $H^l_Y(F^l_{enc}(G(x)))$, the $G(x)$ re-embedded with $F:Y{\rightarrow}X$. We checked whether better results can be obtained by changing the coefficient such as $\lambda_{NCE,X}$ when the direction in the direction or shape of a lane can be observed.\\

\begin{align} &L_{DCLGAN}(G,F,D_X,D_Y,H_X,H_Y)\nonumber\\
= &\lambda_{GAN}(L_{GAN}(G,D_Y,X,Y)\nonumber\\
+&L_{GAN}(F,D_X,X,Y)\nonumber\\
+&\lambda_{NCE}L_{PatchNCE_X}(G,H_X,H_Y,X)\nonumber\\
+&\lambda_{NCE}L_{PatchNCE_Y}(F,H_X,H_Y,Y)\nonumber\\
+&\lambda_{idt}L_{identity}(G,F).\label{eq3}\end{align}
\begin{align} &L_{SimDCL}(G,F,D_X,D_Y,H_X,H_Y)\nonumber\\
= &\lambda_{GAN}(L_{GAN}(G,D_Y,X,Y)\nonumber\\
+&L_{GAN}(F,D_X,X,Y)\nonumber\\
+&\lambda_{NCE}L_{PatchNCE_X}(G,H_X,H_Y,X)\nonumber\\
+&\lambda_{NCE}L_{PatchNCE_Y}(F,H_X,H_Y,Y)\nonumber\\
+&\lambda_{sim}L_{sim}(G,F,H_X,H_Y,H_1,H_2,H_3,H_4)\nonumber\\
+&\lambda_{idt}L_{identity}(G,F).\label{eq4}\end{align}
\begin{align} &L_{PatchNCE_X}(G,H_X,H_Y,X)\nonumber\\
=&E_{x\sim X}\sum^L_{l=1}\sum^{S_l}_{s=1}l(\widehat{z^s_l}, z^s_l, z^{S\setminus s}_l)
\label{eq5}\end{align}

\Figure[b](topskip=0pt, botskip=0pt, midskip=0pt)[width=2.5in]{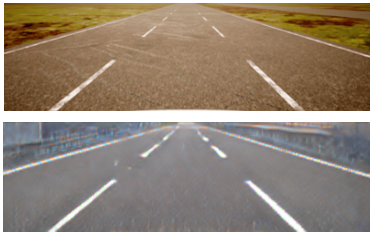}{  A Cw/oD image cropped to 808x245 size and a CwD image generated therefrom for FSIM measurement. \label{fig9}} 

\tab In this case, FSIM \cite{b18} was used to evaluate the degree of lane distortion of the generated image in the case of each adjusted hyperparameter. The edge map similarity between the original image and the fake image is measured from the phase congruency map extracted based on frequency information and the gradient magnitude map extracted based on the gradient of pixel values. To compare the edge information of only lanes, as shown in Fig. \ref{fig9}, 808x245 sized images excluding the background and car bonnet were cropped from each 808x620 image of the CARLA test imageset, and the FSIM of each pair of the Cw/oD image and the CwD image was measured. A model with proper $\lambda_{NCE,X}$ that maintains the original lane shape or direction well will have a high FSIM value. The weight determined in this way is then used for FID measurement. The relationship and differences between FSIM and FID are shown in Fig. \ref{fig10}. FSIM can directly measure the similarity between two paired images, but since FID compares the distribution of images, it is necessary to compare imagesets when measuring FID.

\tab TABLE \ref{table5} shows the average FSIM value in each $\lambda_{NCE,X}$. In all cases, the weight that seems to have the least lane distortion among the weights trained more than 15 times was selected, and 611 images of the test set of the CARLA domain were used. For FSIM measurements, quality\_metrics. FSIM of the image\_similarity\_measures Python package was used \cite{b18}. $\lambda_{NCE,X}$ value 3, which has the highest FSIM value 0.451 and is judged to have the least distortion, was determined as the optimal condition.

\begin{table}[h]
\setlength{\tabcolsep}{13.5pt}
{\tabulinesep=2mm
\begin{tabu}{c|cccc}
\hline
$\lambda_{NCE,X}$ & 1 & 2 & 3 & 4\\
\hline
FSIM & 0.439 & 0.423 & 0.451 & 0.403 \\
\hline
\end{tabu}}
\caption{Mean FSIM values (611 pairs of paired Cw/oD images \& CwD images) for each $\lambda_{NCE,X}$.}
\label{table5}
\end{table}

\Figure[h](topskip=0pt, botskip=0pt, midskip=0pt)[width=3.3in]{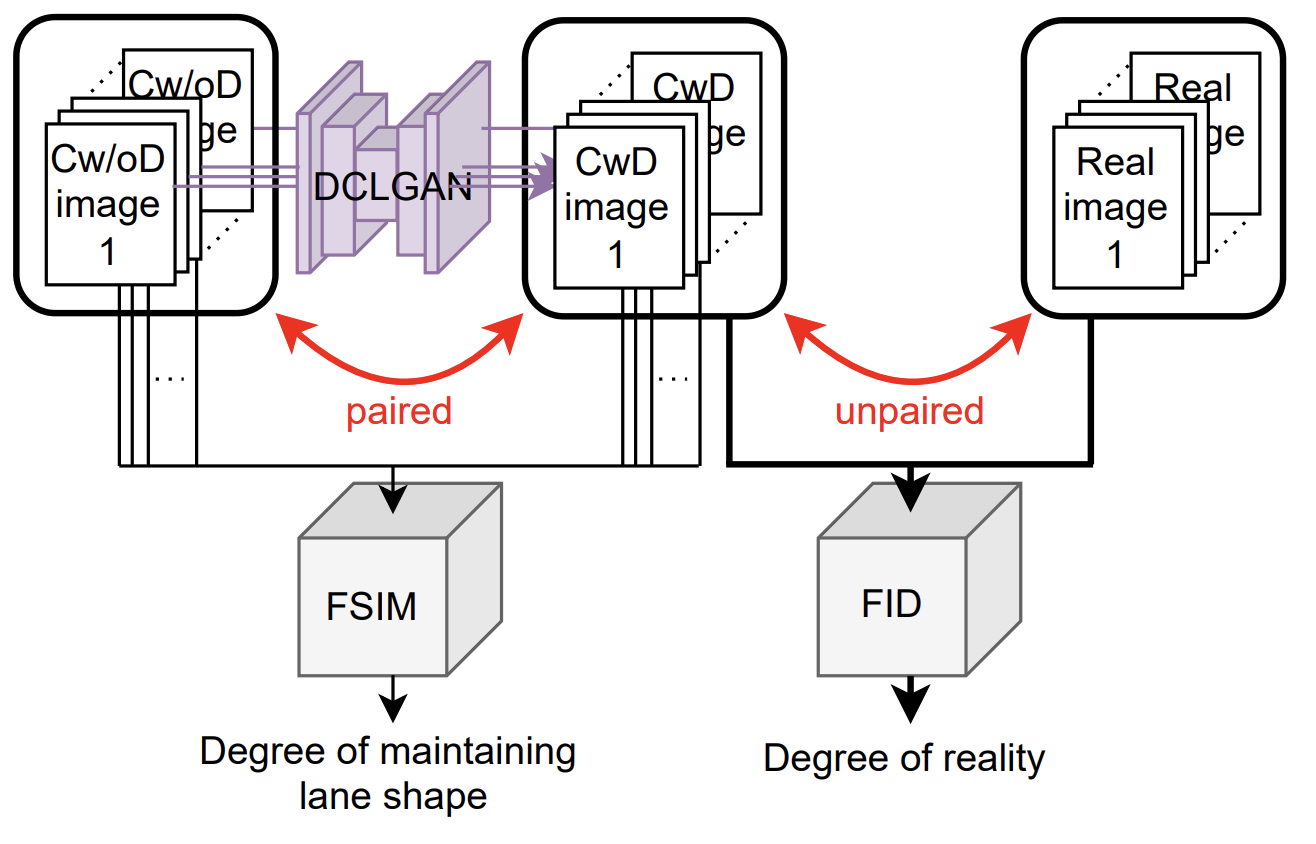}{  A diagram of the relationship between Cw/oD  images, CwD images, and real-world images with measuring FSIM and FID from them. When a CwD image is generated from each CARLA image, the FSIM is measured from each pair. FID is measured between unpaired CwD imageset and real-world imageset to measure how realistic the generated images are. \label{fig10}} 

\subsection{Extracting Ground Truth from Lane Segmentation Images}
\tab To obtain CARLA’s ground truth data automatically, we performed several works. Fig. \ref{fig11} shows this process. First, we created the same map except for the shape of the line. This map's line was fully connected shown as Fig. \ref{fig11}(a). Then, the vehicle ran in the line-connected map while recording its movement and collecting segmentation images. Second, the recording was turned on in the original map, and saved the test images. Then, We found the specific pixel color belonging to the line and counted the number of points for each h\_sample which is the height value that lanes can exist. Lastly,  line order was designated.\\
\tab This algorithm was implemented as described in Algorithm \ref{alg1} only for the second road. When searching based on height, the number of possible points in one height is 2, 3, and 4 
\Figure[t!](topskip=0pt, botskip=0pt, midskip=0pt)[width=3.3in]{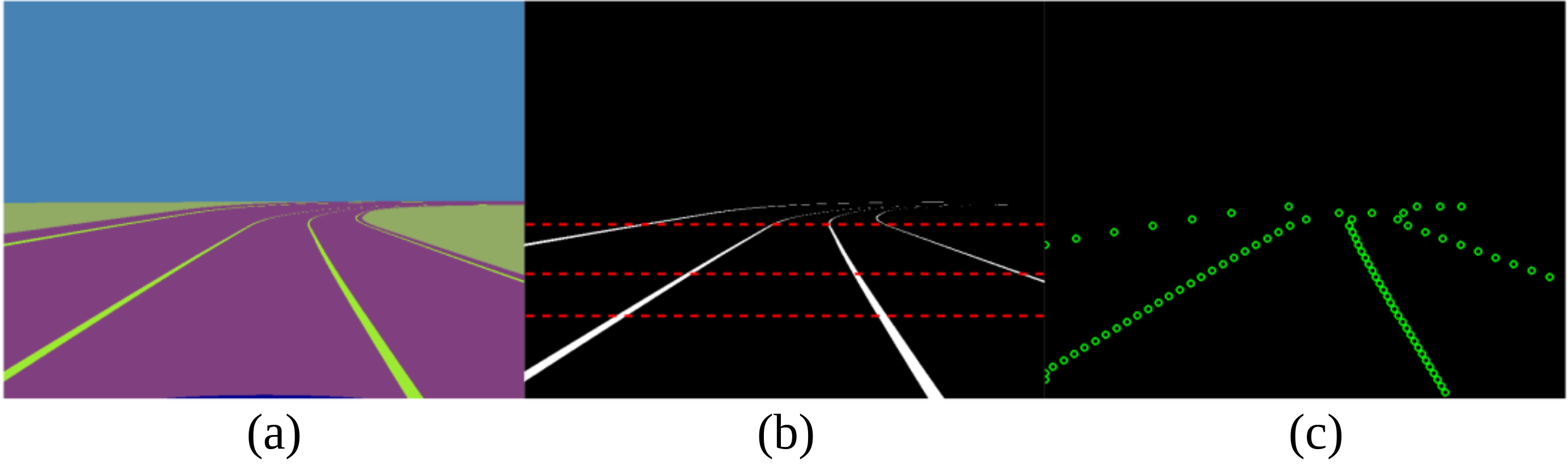}{The segmentation image from line connected map (a), extracting the specific pixel color (b), and ground truth points (b). Red lines in (b) show the example of each case for Algorithm 1.\label{fig11}} 
\begin{algorithm}[t]
\caption{Automatically Extracting Ground Truth}\label{euclid}
\hspace*{\algorithmicindent} \textbf{Input} Segmentation Images\\
\hspace*{\algorithmicindent} \textbf{Output} Ground Truth Data
\begin{algorithmic}[1]
\State $h\_sample \gets $ the list of height values that lines can exist 
\State $line\_rgb \gets $ the RGB value of segmented line
\State $w \gets $ the width of images
\State $GT \gets $ the 2D array of ground truth data. Row corresponds to h\_sample and column corresponds to line number
\For{$i = h\_sample$}
\For{$j = 1,2,...,w$}
\State Put $w$ in the $GT[i]$ that\\ \tab[1cm]satisfies \textbf{Input}${[i][j] = line\_rgb}$ 
\EndFor
\If{$len(GT[i]) = 2$}
\State Set line 2, 3
\Else
\If{$len(GT[i]) = 3$}
\If{$GT[i][1] < w/2$}
\State Set line 1, 2, 3
\Else
\State line 2, 3, 4
\EndIf
\Else
\State Set line 1, 2, 3, 4
\EndIf
\EndIf
\EndFor
\end{algorithmic}
\label{alg1}
\end{algorithm}
\Figure[hbt!](topskip=0pt, botskip=0pt, midskip=0pt)[width=3.3in]{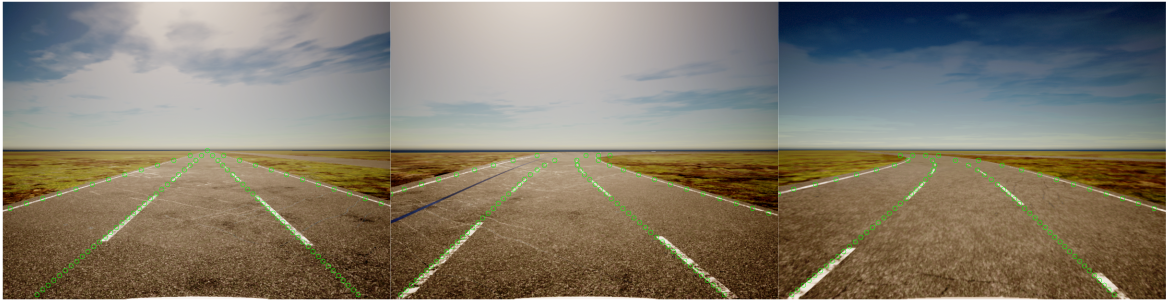}{Results of extracting ground truth (green points) on the original map images.\label{fig12}} 
(top red line of Fig. \ref{fig11}(b)). Two points mean line 2 and line 3 in the left order (bottom red line of Fig. \ref{fig11}(b)). The three points mean line 2, line 3, and line 4 if a second point is on the left based on the middle of the image (middle red line of Fig. \ref{fig11}(b)). Conversely, if the second point is on the right, it means line 1, line 2, and line 3. Lastly, The four points mean line 1, line 2, line 3, and line 4.\\
\section{Experiments and Results}

\tab To show the usefulness of GAN in simulation driving tests, three methods were used to analyze data. First, FID measured the reality of virtual-world images by calculating differences in the density of two distributions. FID showed DCLGAN made CARLA images to realistic. Second, lane segmentation accuracy proved ENet-SAD understood realistic images easier. Third, GPS data showed CwD images made driving results more realistic. We selected two road types; curved roads, and straight roads. Then, the curved road was analyzed by using Root Mean Square Error (RMSE) and the straight road was considered by the ability of lane restoring.

\subsection{FID score}
\tab Cw/oD2Real image gap was obtained by measuring the FID between 5498 CARLA images without DCLGAN and 6929 real-world images, and CwD2Real image gap was obtained by measuring the FID between 5298 images passed through DCLGAN and 6929 real-world images. This relationship is depicted in Fig. \ref{fig13}.
\Figure[hbt!](topskip=0pt, botskip=0pt, midskip=0pt)[width=2.4in]{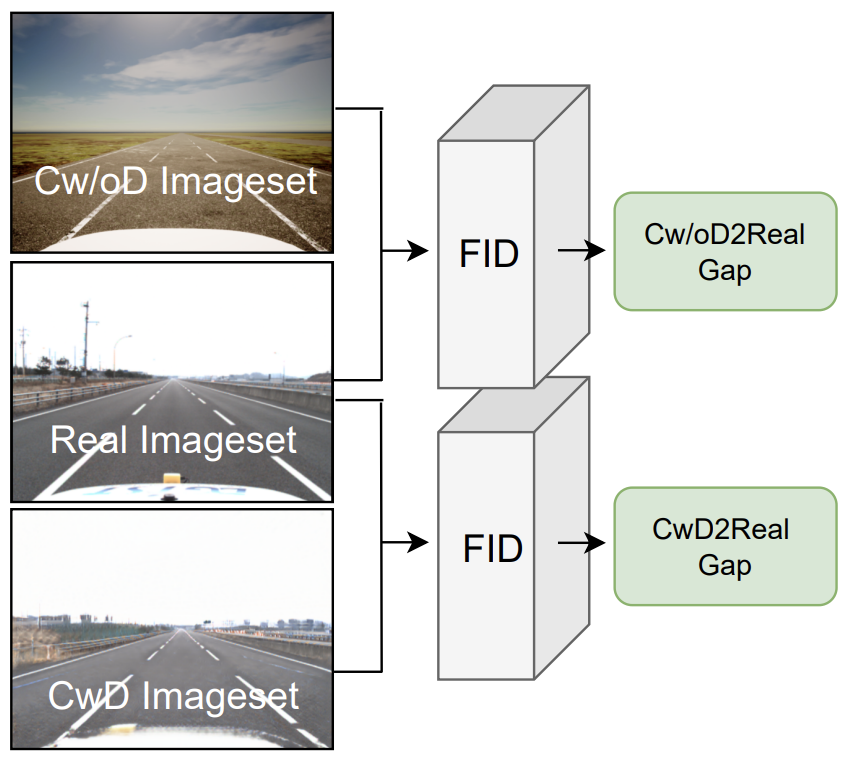}{  Cw/oD2Real gap (Cw/oD vs real-world imageset) and CwD2Real gap (CwD vs real-world imageset) measured by FID \label{fig13} }
\\ \tab As a result of training by setting $\lambda_{NCE,X}$ to 3, the weight(epoch 26) obtained with FSIM value of 0.451 was used. An example of an image translated using this weight is shown in Fig. \ref{fig14}. The lanes of CARLA are inherently blurry and indistinguishable from the surrounding roads, whereas the lanes of the real-world domain are clear and distinctly visible in the distance. This DCLGAN weight translates the image by well reflecting the difference in lane characteristics between these domains without distorting the shape of the lane.\\
\tab TABLE \ref{table6} is the FID measurement of the CARLA images before and after conversion using the corresponding weights. Considering the lack of diversity in our dataset, it was also measured when the dimensionality of the feature was 64, 192, 768, not only 2048. In the case of CARLA domain, 5498 images that were used as trainsets were  used, and in the case of real-world domain, 6929 images were used in the same way. Additionally, 4585 separate images were used for Real vs Real comparison. In all cases, the FID value of CwD was lower than that of Cw/oD. (7.34 vs 1.20, 24.10 vs 4.67, 2.05 vs 1.17, 321.97 vs 240.80) That is, as we saw earlier, DCLGAN changed the CARLA image characteristics such as road color and lane clarity to fit the real-world domain, and it was possible to quantitatively confirm the reduction of the gap with FID.

\Figure[h](topskip=0pt, botskip=0pt, midskip=0pt)[width=2.9in]{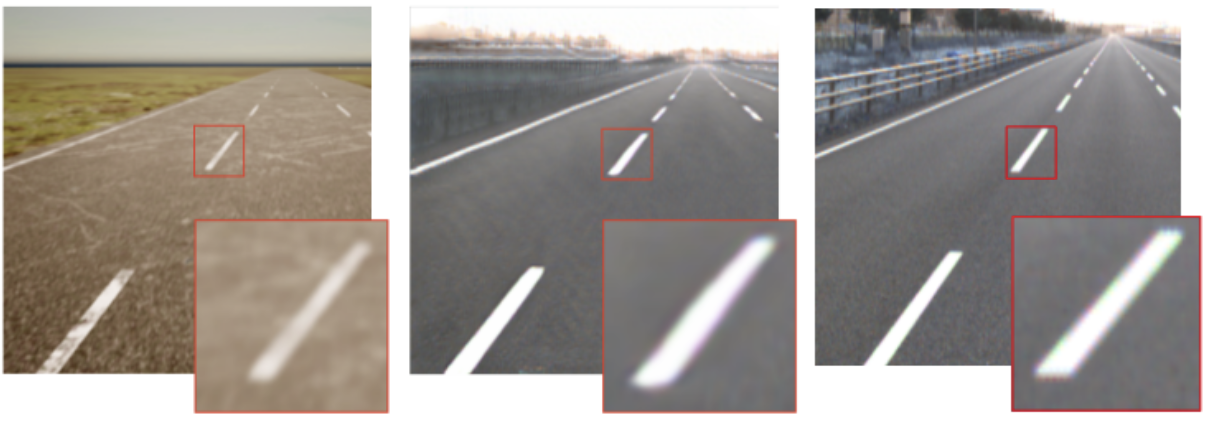}{  From left, Cw/oD, paired CwD, and real-world lane image for comparison. The lane on the real-world domain side is clearer, it can be easily distinguished from the surroundings, and the far side can be seen well. \label{fig14} }
\begin{table}[h]
\setlength{\tabcolsep}{4pt}
{\tabulinesep=2mm
\begin{tabu}{c||ccc}
\hline
Feature Dimensionality & Real vs Cw/oD & Real vs CwD & Real vs Real \\
\hline\hline
64 & 7.34 & 1.20 & 0.01 \\
\hline
192 & 24.10 & 4.67 & 0.05\\
\hline
768 & 2.05& 1.17 & 0.15 \\
\hline
2048 & 321.97 & 240.80 & 86.50 \\
\hline
\end{tabu}}
\caption{Real vs Cw/oD, Real vs CwD, Real vs Real FID values for each dimensionality of the feature. The FID values of the CwD images subjected to DCLGAN were always smaller.}
\label{table6}
\end{table}
\subsection{Lane Segmentation Accuracy}
\begin{table}[h]
\setlength{\tabcolsep}{18pt}
{\tabulinesep=3mm
\begin{tabu}{cccc}
\hline
 &Cw/oD&CwD&Real\\
\hline\hline
Accuracy(\%) & 82.85 & 84.94 & 89.73\\
\hline
\end{tabu}}
\caption{Cw/oD, CwD, Real-world Lane segmentation accuracy.}
\label{table7}
\end{table}
\tab Lane segmentation accuracy in each domain was evaluated followed by the TuSimple evaluation method. Prediction results were obtained using ENet-SAD used for LKAS mentioned 3.A.2. TABLE \ref{table7}. shows the specific results and Fig. \ref{fig15} describes specific flow.\\
\tab This result means that CwD has become similar to the image of reality. This is because DCLGAN corrected the color of CARLA's roads and incomplete cracked lanes. Moreover, DCLGAN modified CARLA images looks like train image dataset from real-world. This made Enet-SAD easier to understand.

\Figure[h](topskip=0pt, botskip=0pt, midskip=0pt)[width=3.3in]{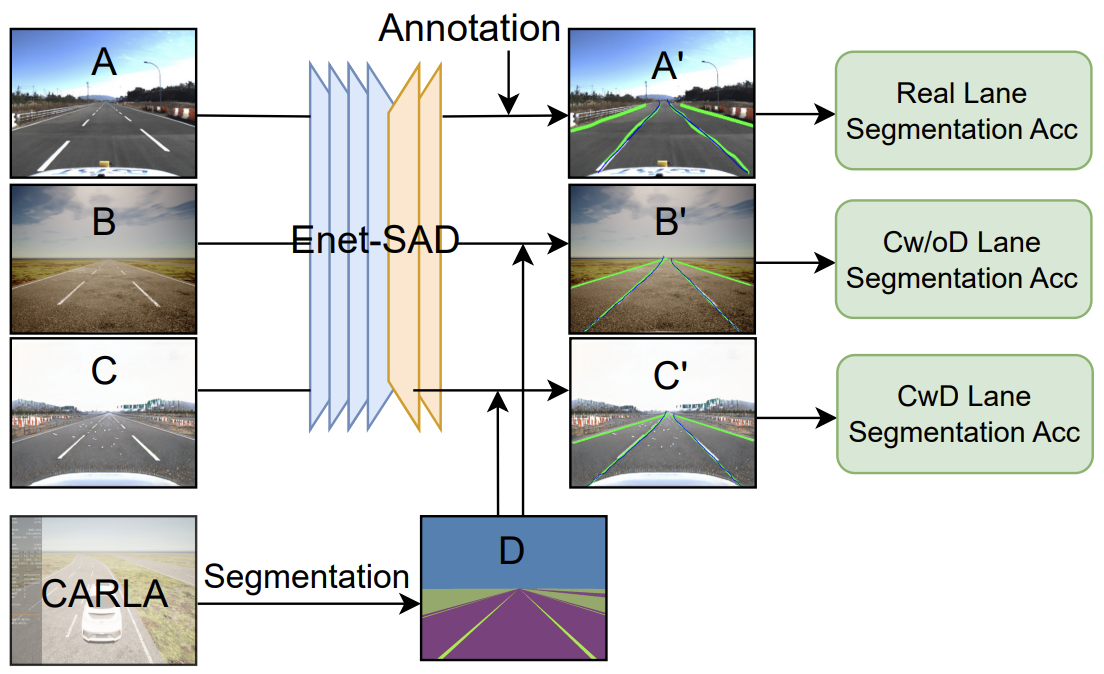}{ Prediction results were got from Enet-SAD. Then accuracy was calculated followed by TuSimple evaluation method.  \label{fig15}}

\subsection{Difference of GPS Orbit}
 \tab To compare the driving test results, we collected GPS data and analyzed them by calculating RMSE. For data collection, The LKAS algorithm was tested in CARLA. The vehicle was spawned and turned on LKAS. The starting position was (East, North) = (44549054.51, 394557162.27) as UTM. After finishing a lap, we repeated this 20 times. Fig. \ref{fig16} shows the overall setting for GPS and Fig. \ref{fig17} shows the overall driving orbital in each domain. After data collection, to utilize RMSE, we selected two road types (slightly curved road, and steep curved road) and two sections for each road type. In Fig. \ref{fig17}, (a), (b), (c), and (d) described each section. Then, The RMSE between the ideal path connecting the central points of the road and each driving path was calculated. More specifically, there are two ways to calculate RMSE: distance east (x) and distance north (y). Table N shows the mean RMSE values of 10 experiments of CwD and Cw/oD for each section.
 \Figure[h](topskip=0pt, botskip=0pt, midskip=0pt)[width=3.3in]{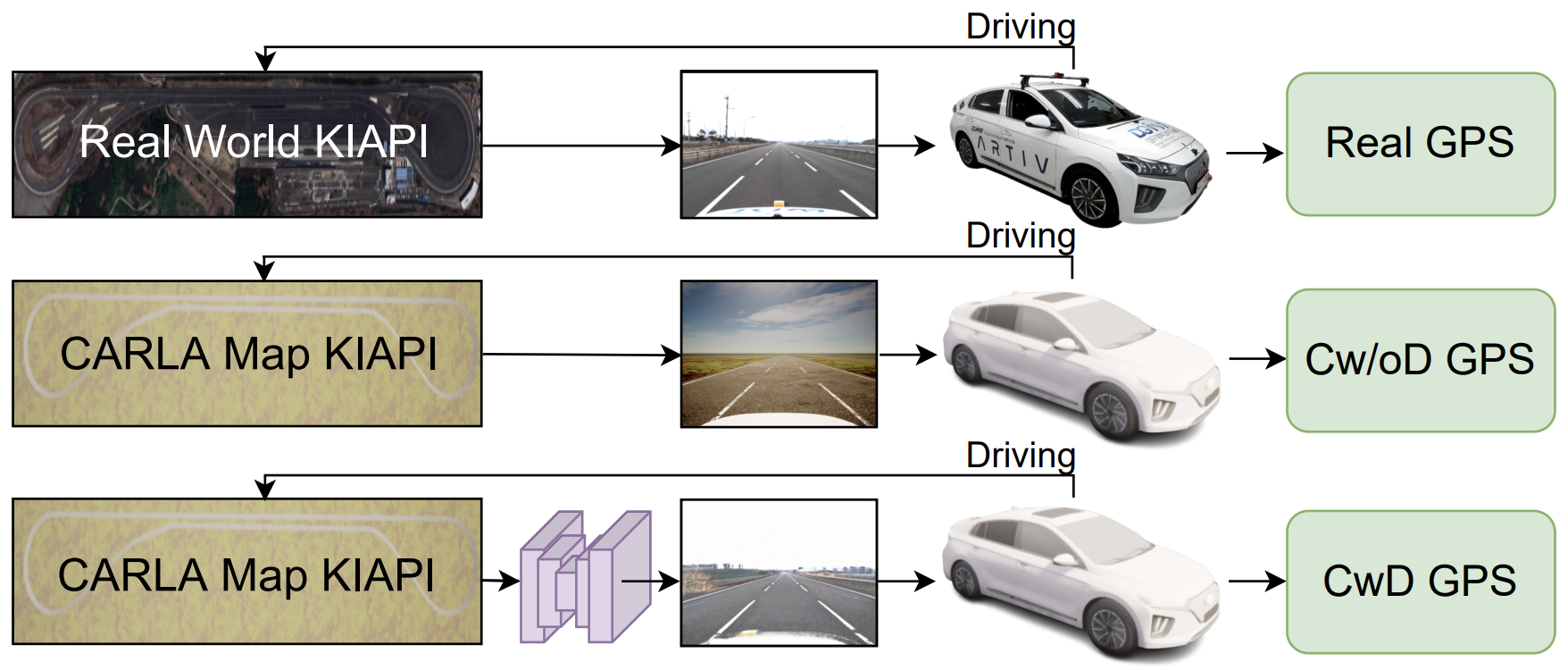}{GPS setting and method flow for each domain.\label{fig16}} 
 
 \tab As shown in TABLE \ref{table8}, CwD ran closer to the center route than Cw/oD. According to section 1 and section 4, the steep curved roads, CwD was more stable. However, in section 2 and section 3 which are slightly curved roads, CwD and Cw/oD were almost the same. Moreover, CwD finished whole laps but Cw/oD failed two times. These results mean that Cw/oD can cause bad driving results because of lane segmentation accuracy, especially on the steep curved road which is the most important part of the map.
 \Figure[hbt!](topskip=0pt, botskip=0pt, midskip=0pt)[width=6.6in]{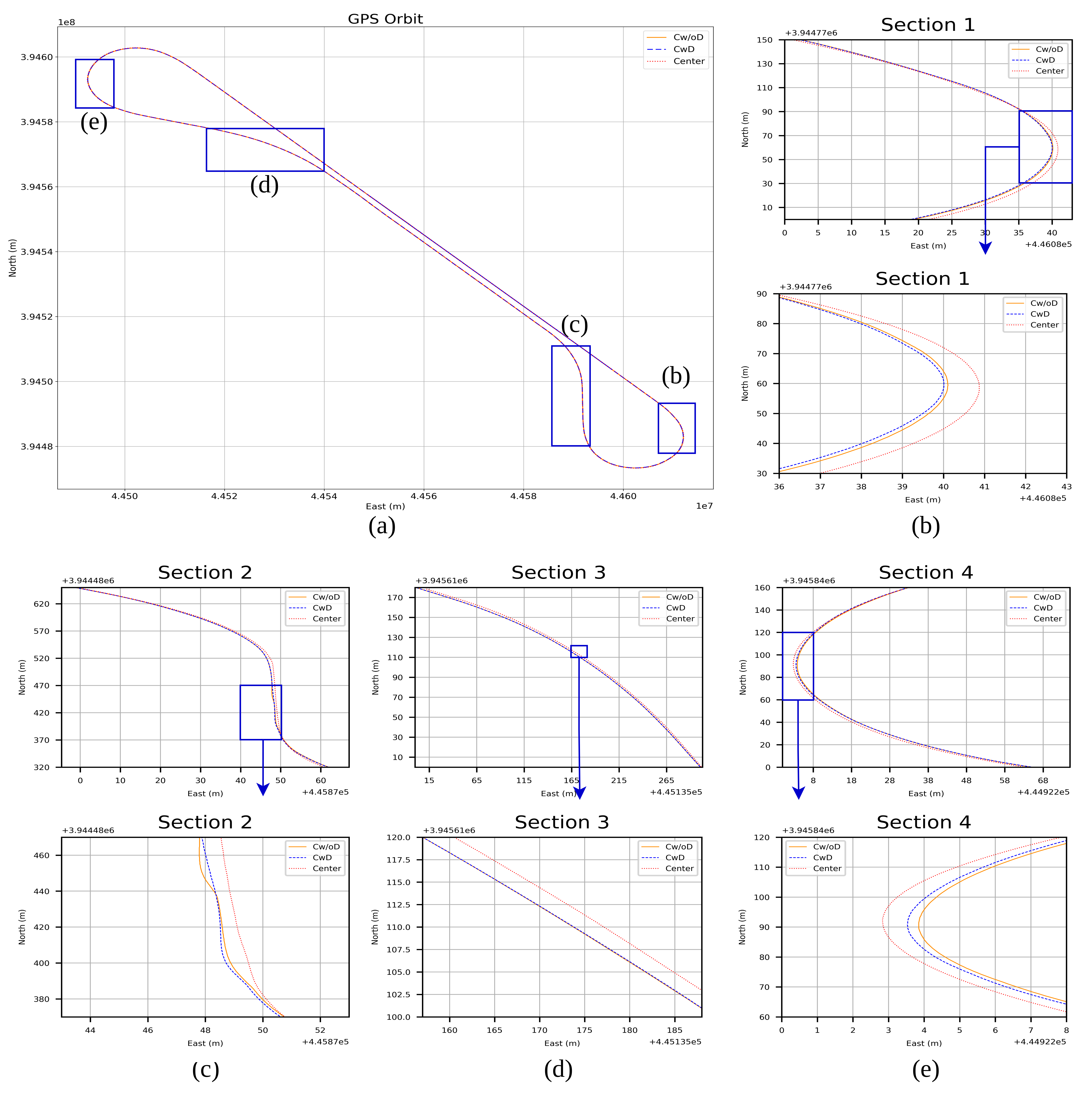}{ GPS orbit of CwD, Cw/oD, and center path. (a) Position of each section for RMSE calculation. (b), (c), (d), (e) Magnified view of each section. \label{fig17}} 
\begin{table*}[t!]
\setlength{\tabcolsep}{12pt}
\centering
{\tabulinesep=2.5mm
\begin{tabular}{c|cc|cc|cc|cc|c}
\hline
{  }     & \multicolumn{2}{c|}{{  Section1}}                             & \multicolumn{2}{c|}{{  Section 2}}                             & \multicolumn{2}{c|}{{  Section 3}}                            & \multicolumn{2}{c|}{{  Section 4}}                            & {  }             \\ \hline
{  }     & \multicolumn{1}{c}{{   x (m)}}     & {  y (m)}     & \multicolumn{1}{c}{{   x (m)}}     & {  y (m)}      & \multicolumn{1}{c}{{   x (m)}}     & {  y (m)}     & \multicolumn{1}{c}{{   x (m)}}     & {  y (m)}     & {  Success Rate ($\%$)} \\ \hline\hline
{  CwD}  & \multicolumn{1}{c}{{  0.753}} & {  1.941} & \multicolumn{1}{c}{{  0.232}} & {  1.579}  & \multicolumn{1}{c}{{  0.182}} & {  0.111} & \multicolumn{1}{c}{{  0.550}} & {  0.998} & {  100}          \\ \hline
{  Cw/oD} & \multicolumn{1}{c}{{  1.115}} & {  2.099} & \multicolumn{1}{c}{{  0.252}} & {  1.574}  & \multicolumn{1}{c}{{  0.192}} & {  0.118} & \multicolumn{1}{c}{{  0.719}} & {  1.449} & {  71.4}           \\ \hline
{  Cw/oD - CwD} & \multicolumn{1}{c}{{  0.337}} & {  0.158} & \multicolumn{1}{c}{{  0.021}} & {  -0.055} & \multicolumn{1}{c}{{  0.003}} & {  0.001} & \multicolumn{1}{c}{{  0.139}} & {  0.481} & {  }             \\ \hline
\end{tabular}}

\caption{The degree to which CwD and Cw/oD deviate from the center of the road in each session. CwD drove a complete lap in all cases, showing a 100\% success rate. Cw/oD was 80\%.}
\label{table8}
\end{table*}

\subsection{LANE RESTORING}

\tab In this section, we tried to evaluate how well a vehicle can return when it veers out of its lane. If CwD recognizes the lane better than Cw/oD, the ability to return to the existing lane within reasonable time when it deviates from the lane should be better. To confirm this, a lane-keeping driving test was conducted after locating the car at +-90cm, 120cm, 130cm, 140cm, and 150cm in the direction perpendicular to the lane from the common driving start GNSS location (35.6524837116667, 128.397828661667). (+ is right, - is left) In common, if the vehicle starts from its own location and returns to the original lane within 2 minutes and continues to keep the lane stably, it was judged as a success.
\tab The results are shown in Fig. \ref{fig18} and TABLE \ref{table9}. As shown in TABLE \ref{table9}, Cw/oD failed when starting by moving 1.5 m, 1.4 m to the right, and 1.2 m, 1.3 m, 1.4 m, and 1.5 m to the left. However, CwD succeeded in all but the right 1.5m, left 1.3m, 1.4m, and 1.5m, and as shown in Fig. \ref{fig18}, the path returning to the original lane was confirmed even under relatively extreme conditions such as +1.4m and -1.3m. Compared to Cw/oD, it can be seen that CwD has better restoring force to return to the center of the lane by better recognizing the lane even when it is above the lane away from the center of the lane.

 \Figure[!t](topskip=0pt, botskip=0pt, midskip=0pt)[width=3.3in]{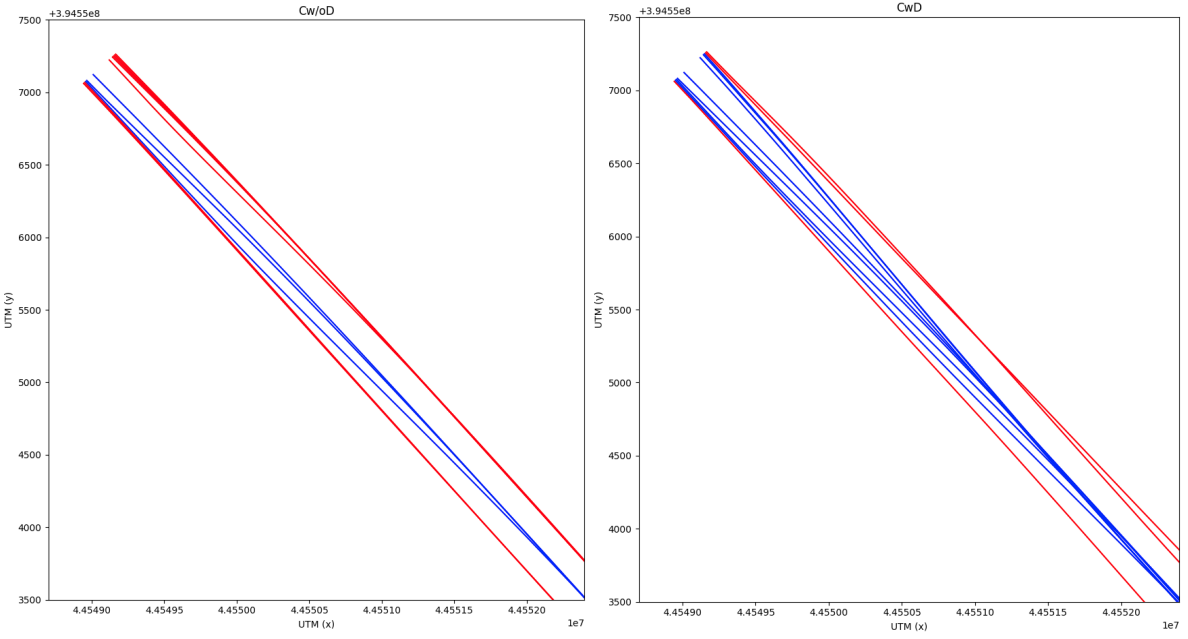}{ Lane restoring GPS path; Red (X), Blue (O). \label{fig18}}

\begin{table}[h]
\setlength{\tabcolsep}{4pt}
{\tabulinesep=1.2mm
\begin{tabu}{c||cccccccccc}
\hline
Distance (m)&1.5&1.4&1.3&1.2&0.9&-0.9&-1.2&-1.3&-1.4&-1.5\\
\hline
Cw/oD & X & X& O& O& O& O&X&X& X& X\\
\hline
CwD & X & O& O& O& O& O& O& O&X& X\\
\hline
\end{tabu}}
\caption{Lane return success or not according to the distance from the center of the lane.}
\label{table9}
\end{table}
 \tab 

\section{Conclusion}
\tab This paper proposed the issues about unrealistic images when the algorithm was tested in simulation and the solution of this problem by utilizing DCLGAN model. The images extracted from CARLA's camera were converted to the real domain. Then, FID quantitatively showed that it became more similar to reality, especially in our KIAPI high-speed circulation environment.  through the Lane segmentation accuracy, we showed that the image of DCLGAN is similar to the reality, of our train dataset, so that Enet-SAD had a better understanding of the lane. Finally, we evaluated the difference in the overall trajectory with the difference in GPS orbitals as area and compared the driving trajectory of critical parts through RMSE (x) and RMSE (y). These indicated that CwD is more similar to the driving result in reality. Future work will concentrate on applying AEB and SCC and implementing dynamic models more precisely through CARSIM.

\EOD

\end{document}